\documentclass[
]{ceurart}

\usepackage{listings}
\usepackage{comment}
\usepackage{multirow}
\usepackage{multicol}
\usepackage{url}
\usepackage{pifont}
\begin{document}

\copyrightyear{2026}
\copyrightclause{Copyright for this paper by its authors.
  Use permitted under Creative Commons License Attribution 4.0
  International (CC BY 4.0).}

\conference{WOP at ISWC’26: 17th Workshop on Ontology Design and Patterns at the International Semantic Web Conference, Bari, Italy}


\title{When Does Bigger Help? A Controlled Study of LLM Scale for Ontology Learning}

\author[1]{Hamed {Babaei Giglou}}[%
orcid=0000-0003-3758-1454,
email=hamed.babaei@tib.eu,
]
\address[1]{TIB Leibniz Information Centre for Science and Technology, Hannover, Germany}

\author[1,2]{Sören Auer}[%
orcid=0000-0002-0698-2864,
email=auer@tib.eu,
]
\address[2]{L3S Research Center, Leibniz University of Hannover, Hannover, Germany}

\author[1]{Jennifer D'Souza}[%
orcid=0000-0002-6616-9509,
email=jennifer.dsouza@tib.eu,
]

\begin{abstract}
The effect of Large Language Model (LLM) scale on ontology learning (OL) performance remains insufficiently characterized. We present a controlled evaluation of 13 models spanning dense and Mixture-of-Experts variants from the Qwen3.5 and Qwen3.6 lineages, together with proprietary GPT release variants, using the OntoLearner retrieval-augmented generation pipeline. All models are evaluated with the same embedding model, retrieval configuration, prompt templates, decoding settings, datasets, and metrics on term typing, taxonomy discovery, and non-taxonomic relationship extraction across four biomedical and materials science and engineering ontologies. Within the dense Qwen3.5 lineage, increasing parameter count primarily improves precision rather than recall, with the largest gains occurring between 9B and 27B parameters. However, the effect of scale is neither monotonic nor uniform across tasks and domains. Dense 27B models outperform substantially larger sparse models on term typing, whereas larger Mixture-of-Experts models achieve the strongest open-weight results on taxonomy discovery. Non-taxonomic relationship extraction remains difficult across model scales, particularly for the Materials Data Science ontology. Performance differences across matched Qwen variants and proprietary GPT releases further indicate that architecture and model lineage can outweigh nominal parameter count. These findings show that model size alone is an insufficient selection criterion for OL and provide empirical guidance for reproducible LLM-assisted ontology engineering.
\end{abstract}

\maketitle

\begin{itemize}[]
\item \textbf{URL}: \url{https://ontolearner.readthedocs.io/}
\item \textbf{GitHub}: \url{https://github.com/sciknoworg/OntoLearner}
\item \textbf{PyPi}:  \url{https://pypi.org/project/OntoLearner/}
\item \textbf{Huggingface}:  \url{https://huggingface.co/collections/SciKnowOrg/ontolearner-benchmarking}
\item \textbf{License}: MIT License
\item \textbf{Experimental Resources:} \url{https://doi.org/10.5281/zenodo.21719027}
\end{itemize}

\section{Introduction}
Scientific research is becoming increasingly data- and automation-intensive, particularly in biomedicine and materials science, where high-content screening, automated experimentation, and self-driving laboratories generate growing volumes of heterogeneous data, experimental procedures, and domain terminology \cite{bock2022high,tom2024self}. For these outputs to be reused across laboratories, databases, and computational systems, they must be represented in forms that are not only digitally accessible but also findable, interoperable, and machine-actionable \cite{wilkinson2016fair}. Structured knowledge representations are therefore essential for semantic search, cross-source integration, reproducible analysis, scientific reasoning, and AI-assisted discovery \cite{auer2025open,wang2023scientific,eger2025transforming}. Ontologies provide the formal conceptual layer needed to organize domain entities and relationships consistently across heterogeneous sources. However, constructing and maintaining high-quality ontologies remains a labor-intensive and expert-driven process, limiting the scalability, adaptability, and reproducibility of ontology engineering \cite{kendall2019ontology}.

Recent advances in Large Language Models (LLMs) present new opportunities for automating ontology learning (OL) tasks \cite{giglou2023llms4ol}. Given lexical terms and ontology-grounded context, LLMs can support assigning terms to semantic types, identifying taxonomic relationships between concepts, and predicting non-taxonomic relations. Their ability to perform these tasks with little or no task-specific training makes LLM-based OL a promising approach for reducing the manual effort involved in ontology construction, extension, and maintenance, and for supporting the scalable development of knowledge graphs across scientific domains.
The first LLMs4OL study established an empirical foundation for investigating LLMs in OL by evaluating a broad selection of model families across term typing, taxonomy discovery, and non-taxonomic relation extraction \cite{giglou2023llms4ol}. Its primary objective was to determine whether general-purpose and domain-specific language models could perform these tasks at all. Accordingly, it compared models with different architectures, training objectives, domain specializations, and parameter counts, including BERT~\cite{bert}, BART~\cite{bart}, Flan-T5~\cite{longpre2023flan}, BLOOM~\cite{BLOOM}, LLaMA~\cite{llama}, GPT~\cite{dale2021gpt}, and PubMedBERT~\cite{gu2021domain}. Comparisons between selected Flan-T5 and BLOOM variants provided initial indications that larger models could sometimes perform better, while task-specific fine-tuning showed that smaller models could also outperform substantially larger ones. Model size was therefore not isolated from model family, architecture, training strategy, or domain adaptation.

The LLMs4OL Challenges, organized at the International Semantic Web Conference (ISWC), subsequently broadened this investigation through shared tasks, datasets, evaluation phases, and metrics for community-developed OL systems \cite{BabaeiGiglou2024LLMs4OL,BabaeiGiglou2025LLMs4OL}. In the 2024 challenge, participants explored fine-tuning and knowledge-enhanced prompt-tuning \cite{RWTHDBIS2024,SKHNLP2024}, prompt-tuning across multiple LLMs \cite{TheGhost2024}, retrieval-augmented generation \cite{Phoenixes2024}, hybrid LLM--rule-based processing \cite{TSOTSALearning2024}, and conventional machine-learning methods combined with LLMs \cite{SilpNLP2024}. The 2025 challenge expanded this methodological range through systems combining prompt engineering and embeddings \cite{SBUNLP2025}, RAG, few-shot prompting, embedding ensembles, and lightweight fine-tuning \cite{Alexbek2025}, clustering and retrieval-augmented extraction \cite{SilpNLP2025}, data augmentation and candidate filtering \cite{IRIS2025}, and multi-model deliberation \cite{DREAMLLMs2025}. Some submissions also compared differently sized models \cite{Phoenixes2025,ELLMO2025}, but these comparisons generally varied model family, architecture, adaptation strategy, or system configuration alongside size. They were consequently suited to identifying effective OL systems, but not to estimating the independent effect of parameter scale. This leaves a fundamental question unanswered:

This gap reflects a broader and still unresolved debate in foundation-model research concerning whether larger LLMs are systematically more capable. Early neural scaling-law studies identified smooth power-law reductions in pretraining loss as model parameters, training data, and computational resources increased \cite{kaplan2020scaling}. Subsequent compute-optimal analyses, however, demonstrated that parameter count alone is an incomplete indicator of capability: a smaller model trained on substantially more data can outperform a much larger but under-trained model \cite{hoffmann2022training}. Model scale must therefore be understood in relation to training data, compute, and architecture rather than through parameter count alone. The relationship becomes even less predictable at the downstream-task level. Some capabilities have been reported to appear only after models exceed particular scale thresholds \cite{wei2022emergent}, whereas other work has shown that apparently discontinuous emergence may partly result from the choice of evaluation metric \cite{schaeffer2023mirage}. Inverse and U-shaped scaling have also been observed, whereby performance can plateau, deteriorate, or change direction as models grow \cite{mckenzie2023inverse,wei2023inverse}. A recent re-examination similarly found smooth and predictable scaling in only a minority of downstream-task settings \cite{lourie2025scaling}. Improvements in pretraining objectives or aggregate benchmarks therefore cannot be assumed to transfer monotonically to specialized knowledge-engineering tasks.

This uncertainty is especially consequential for OL because its constituent tasks impose different semantic and structural demands. Term typing requires calibrated selection among candidate ontology classes, taxonomy discovery requires reasoning over hierarchical relations, and non-taxonomic relationship extraction requires distinguishing heterogeneous domain relations. Increasing scale may therefore affect precision, recall, and relational reasoning differently across tasks and domains. Furthermore, ``model size'' is not entirely unambiguous across architectures. Dense models activate their full parameter set, whereas Mixture-of-Experts (MoE) models separate total model capacity from the smaller number of parameters activated for an individual input \cite{fedus2022switch}. Architecture, training composition, instruction tuning, and alignment may consequently mediate or outweigh the effect of nominal parameter count.

Although existing OL studies indicate that model choice and capacity matter, they do not isolate these factors through a sufficiently broad within-family scale sweep under fixed experimental conditions. We therefore investigate the following research questions:

\begin{description}
\item[\textbf{RQ1: Within-family scaling.}] How does increasing parameter count within a common dense model family affect precision, recall, and F1 performance in OL, and are the resulting improvements monotonic?

\item[\textbf{RQ2: Task and domain sensitivity.}] How does the effect of model scale vary across OL tasks and across scientific domains with different conceptual structures?

\item[\textbf{RQ3: Architecture and model lineage.}] How do dense versus MoE architectures and differences across Qwen and GPT release lineages affect OL performance beyond nominal parameter count?
\end{description}

We address these questions through a controlled benchmark study using OntoLearner \cite{giglou2026ontolearnermodularpythonlibrary}, a reproducible ecosystem for LLM-based OL. We evaluate 13 models spanning the Qwen3.5~\cite{qwen3.5}, Qwen3.6~\cite{qwen3.6-27b,qwen36_35b_a3b}, and proprietary GPT lineages. The Qwen3.5 models enable a within-family dense-model scale sweep from 0.8B to 27B parameters, complemented by larger sparse MoE variants with 35B total and 3B active parameters and 122B total and 10B active parameters. The Qwen3.6 models permit matched-scale comparisons with Qwen3.5 across a 27B dense model and a 35B-A3B MoE model, while GPT-5.5~\cite{openai_gpt55_api_2026} and the GPT-5.6 Luna~\cite{openai_gpt56_luna_2026}, Terra~\cite{openai_gpt56_terra_2026}, and Sol~\cite{openai_gpt56_sol_2026} variants provide a comparison across proprietary model releases. All models are evaluated on term typing, taxonomy discovery, and non-taxonomic relationship extraction over biomedical and materials science and engineering ontologies. By holding the embedding model, retrieval configuration, retrieved context, prompt templates, decoding conditions, datasets, and evaluation metrics constant, we examine the effects of parameter scale, dense versus sparse architecture, model release, task structure, and domain characteristics. The Results section is organized according to the three research questions and provides empirical guidance for model selection in scalable and reproducible ontology engineering.



\section{Preliminaries}
\label{sec:preliminaries}

Following the OL formalization introduced in~\cite{giglou2023llms4ol} and adopted by the LLMs4OL Challenge, we consider three fundamental OL tasks. Let $L$ denote a lexical term extracted from text and $T$ denote a type (i.e., an ontology class). These tasks constitute the basis of our experimental evaluation.

\paragraph{Term Typing.}Term typing aims to assign the most appropriate ontology type to a lexical term. Formally,
\begin{equation}
    OL_{\text{term typing}}(L) := (L, T)
\end{equation}
where the input is a lexical term $L$, and the output is its corresponding ontology type $T$.

\paragraph{Taxonomy Discovery.} Taxonomy discovery identifies hierarchical (\texttt{is-a}) relationships between ontology types. Given two lexical terms, their corresponding ontology types are inferred, and the taxonomic relationship between them is determined. Formally,
\begin{equation}
    OL_{\text{taxonomy discovery}}(L_a, L_b) := (T_a, T_b)
\end{equation}
where $T_a$ and $T_b$ denote the ontology types associated with lexical terms $L_a$ and $L_b$, respectively.

\paragraph{Non-Taxonomic Relationship Extraction.} This task identifies semantic relations other than \texttt{is-a} between ontology types. Given a head lexical term $h$, a relation $r$, and a tail lexical term $t$, the objective is to infer the semantic relation between their corresponding ontology types. Formally,
\begin{equation}
    OL_{\text{non-taxonomic re}}(h, r, t) := (T_h, r, T_t),
\end{equation}
where $T_h$ and $T_t$ are the ontology types corresponding to the head and tail lexical terms, respectively.

\section{Related Work}

OL has long been recognized as a bottleneck-relieving alternative to manual ontology engineering, and several recent surveys examine how this bottleneck manifests in specific application domains. Ivanova and Terzieva~\cite{ivanova2026ontology} systematically review OL techniques for intelligent e-learning environments, mapping classical linguistic, statistical, and logic-based OL methods onto educational data sources and highlighting that LLM-based OL for education remains comparatively under-studied despite the maturity of general-purpose OL research. Their review echoes a broader pattern in the literature: domain adaptation of OL techniques is frequently treated as a secondary concern relative to the development of the underlying extraction methods themselves. 
\paragraph{LLM-Based Ontology Generation Pipelines.} A growing body of work investigates ontology construction directly from natural language using LLMs. Lippolis et al.~\cite{lippolis2025ontologygeneration} introduce two prompting techniques, Memoryless CQbyCQ and Ontogenia, for generating OWL drafts from competency questions and user stories, finding that reasoning-oriented models (e.g., o1-preview) can approach or exceed novice human ontology engineers on structural correctness, while open-weight models like Llama produce more superfluous elements and critical modeling pitfalls. In a related study, Lippolis et al.~\cite{lippolis2025assessing} assess the domain-generalizability of two reasoning-capable LLMs (DeepSeek and o1-preview) across six ontology-engineering projects and find remarkably consistent performance across domains, suggesting that ontology-generation ability is not narrowly tied to domain representation in pre-training data. 
Teplyi and Dosyn~\cite{teplyi2025promptguided} propose a four-stage prompt-guided agent (schema discovery, instance extraction, self-repair validation, ontology alignment) that iteratively enforces domain-range consistency, achieving fact-recall competitive with specialized knowledge graph extraction pipelines such as KGGen while additionally producing schema-consistent output. Their comparison across GPT-4.1-mini, LLaMA-3.3-70b, and Grok-3-mini shows that larger models tend to hypothesize broader schemas that remain partly unpopulated, whereas smaller models produce more compact but fully instantiated ontologies, an early signal of the scale-dependent trade-offs that our work investigates more systematically. Moreover, Fathallah et al.~\cite{fathallah2026extended} extend this line of work with NeOn-GPT, a NeOn-methodology-grounded pipeline incorporating an explicit ontology-reuse stage and automated verification (RDFLib, HermiT/Pellet, OOPS!). Evaluated across four heterogeneous domains using GPT-4o, Mistral, Llama-4, and DeepSeek, they report that all models default to shallow, single-inheritance hierarchies regardless of architecture or parameter count, a ``systematic generative limitation'' rather than a model-specific one, and that entity-level semantic alignment consistently exceeds relation-level alignment. Zhang et al.~\cite{zhang2025olive} similarly combine LLM-driven extraction (via ChatGPT and Llama) with graph-database backends (OLIVE), emphasizing iterative human-in-the-loop correction of hallucinated relations validated through OOPS! and Protégé reasoning.
\paragraph{Fine-Tuning versus Prompting and Model Comparisons.} Several works directly compare model families or adaptation strategies for OL, providing precedent for our size-focused investigation. Doumanas et al.~\cite{doumanas2025finetuning} fine-tune GPT-4 and Mistral 7B on curated ontology-engineering textbook corpora, finding that GPT-4 achieves higher precision and syntactic fidelity at greater computational cost, while Mistral 7B is cheaper and faster but degrades unpredictably across successive fine-tuning rounds, underscoring that model family and scale interact with training strategy in ways that are not yet well characterized. Liu et al.~\cite{liu2026prompt} compare pre-trained prompting, in-context learning, and fine-tuning for casting domain term and relation extraction, finding fine-tuning superior in F1 but noting the substantial annotation cost required to enable it. Cappelli and Di Marzo Serugendo~\cite{cappelli2025methodological} instead rely on a single frontier model (ChatGPT-4o) guided by a structured ``guiding table'' of prompts, reporting high logical consistency but persistently flat, low-attribute-richness taxonomies—again suggesting that prompting strategy alone cannot fully compensate for structural limitations that may be tied to model capacity. None of these studies, however, isolate model size as an independent variable while holding domain, prompting strategy, and evaluation protocol constant.
\paragraph{OL for Knowledge Graphs, Retrieval, and Specialized Domains.} A parallel thread examines OL as an upstream component of larger systems rather than an end in itself. da Cruz et al.~\cite{dacruz2025ontology} show that ontology-guided knowledge graphs—particularly those derived from relational-database schemas—achieve retrieval performance competitive with GraphRAG at substantially lower LLM-inference cost than text-derived alternatives, while Lippolis et al.~\cite{lippolis2026ontoextend} target the complementary problem of requirement-driven ontology extension rather than generation from scratch, showing via expert review that RAG-grounded extension proposals require only minor revision before production use. A similar topic was in the middle of interest to the LLMs4OL challenge community this year~\footnote{\url{https://sites.google.com/view/llms4ol2026/reuse-task}}. Domain-specific deployments further illustrate OL's practical stakes: an ICMHI '25 study~\cite{cherian2025custom} applies OL models to generate custom FHIR extensions for healthcare data interoperability, underscoring how OL quality directly affects downstream system usability in high-stakes settings.

\begin{table}[t]
\centering
\caption{Comparison of representative LLM-based OL studies. \checkmark/\ding{55}\ indicate whether a size was explicitly varied or controlled for.}
\label{tab:related_work_comparison}
\small
\resizebox{\textwidth}{!}{%
\begin{tabular}{@{}p{3cm}p{3.1cm}p{2.3cm}p{2.0cm}c c p{6cm}@{}}
\toprule
\textbf{Study} & \textbf{Model(s)} & \textbf{Approach} & \textbf{Domain(s)} & \textbf{Cross-} & \textbf{Scale} & \textbf{Key Finding} \\
 & & & & \textbf{Domain} & \textbf{Controlled} & \\
\midrule
    Lippolis et al.~\cite{lippolis2025ontologygeneration} & o1-preview, GPT-4, Llama & Prompt (CQbyCQ, Ontogenia) & 10 mixed ontologies & \checkmark & \ding{55} & Reasoning models rival novice engineers; Llama yields more critical pitfalls \\
    Lippolis et al.~\cite{lippolis2025assessing} & DeepSeek, o1-preview & Prompt & 6 domains & \checkmark & \ding{55} & Performance is consistent across domains, not tied to pretraining coverage \\
    Teplyi \& Dosyn~\cite{teplyi2025promptguided} & GPT-4.1-mini, LLaMA-3.3-70B, Grok-3-mini & Agentic + self-repair & General text & \ding{55} & Partial\textsuperscript{$\dagger$} & Larger models hypothesize broader, unpopulated schemas; smaller models produce compact, fully instantiated ones \\
    Fathallah et al.~\cite{fathallah2026extended} & GPT-4o, Mistral, Llama-4, DeepSeek & Prompt + reuse & 4 domains & \checkmark & \ding{55} & All models default to shallow hierarchies regardless of scale or architecture \\
    Zhang et al.~\cite{zhang2025olive} & ChatGPT, Llama & Prompt + HITL & General (KG) & \ding{55} & \ding{55} & Human-in-the-loop correction still required to fix hallucinated relations \\
    Doumanas et al.~\cite{doumanas2025finetuning} & GPT-4, Mistral 7B & Fine-tune & OE textbooks & \ding{55} & Partial\textsuperscript{$\dagger$} & Fine-tuning benefit is model-family-dependent; Mistral degrades across rounds \\
    Liu et al.~\cite{liu2026prompt} &  GPT-4.1-mini & Prompt vs.\ ICL vs.\ FT & Casting & \ding{55} & \ding{55} & Fine-tuning yields best F1 but at substantial annotation cost \\
    Cappelli \& Serugendo~\cite{cappelli2025methodological} & ChatGPT-4o & Prompt (guiding table) & 1 domain & \ding{55} & \ding{55} & High logical consistency but flat, low-attribute-richness taxonomies \\
    da Cruz et al.~\cite{dacruz2025ontology} & LLaMA-3.2-3B & RAG & General & \ding{55} & \ding{55} & DB-derived ontologies rival GraphRAG at lower LLM inference cost \\
    Lippolis et al.~\cite{lippolis2026ontoextend} & GPT-5, o1-preview  & RAG  & General & \ding{55} & \ding{55} & Extension proposals need only minor expert revision before reuse \\
    Cherian et al.~\cite{cherian2025custom} & unspecified & OL for FHIR & Healthcare & \ding{55} & \ding{55} & OL quality directly affects downstream interoperability outcomes \\
\bottomrule
\end{tabular}
}

\vspace{2pt}
\begin{flushleft}
\footnotesize\textsuperscript{$\dagger$}Compares multiple models/families but not multiple sizes \emph{within} a single family under a controlled protocol.
\end{flushleft}

\end{table}

As~\autoref{tab:related_work_comparison} makes clear, existing work has compared different LLM families (GPT vs. Mistral vs. Llama vs. DeepSeek) and, in a few cases, informally noted size-related trends as a byproduct of family comparisons~\cite{teplyi2025promptguided,doumanas2025finetuning}, but no study isolates model scale as an independent variable within a single family while holding domain, prompting strategy, and evaluation protocol fixed. Likewise, cross-domain evaluation is common, but it is rarely paired with a scale sweep: studies that vary domain~\cite{lippolis2025assessing,fathallah2026extended} do not vary size, and the one study that varies size across a handful of models~\cite{teplyi2025promptguided} does not control for domain or family. This leaves open a basic practical question for anyone using LLM-based OL: \textit{does scaling within a model family reliably improve OL quality, and is the marginal gain worth the added computational cost?} We address this gap by using the LLMs4OL task formalization (term typing, taxonomy discovery, non-taxonomic relation extraction) together with the OntoLearner benchmarking ecosystem~\cite{giglou2026ontolearnermodularpythonlibrary} to run a controlled scale sweep across the most recent models from Qwen and GPT model families, evaluated consistently on both biomedical and materials-science ontologies.

\section{Methodology}
\label{sec:methodology}


To evaluate how model parameter scale impacts performance across core OL tasks defined in \autoref{sec:preliminaries}, we adopt a Retrieval-Augmented Generation (RAG) framework adapted from the OntoLearner infrastructure~\cite{giglou2026ontolearnermodularpythonlibrary}~\footnote{See RAG module at OntoLearner website: \url{https://ontolearner.readthedocs.io/learners/rag.html}}. Our pipeline decouples context acquisition from generation by pairing a static dense retriever with the target LLM. This setup isolates the generation capabilities of each LLM scale variant under uniform retrieval conditions. The overall architecture consists of three sequential modules: (1) \textit{Semantic Vector Indexing}, (2) \textit{Similarity-Based Retrieval}, and (3) \textit{Task-Specific LLM Prompting and Generation}. 

\paragraph{Semantic Vector Indexing.} Before running inference, we preprocess the underlying target ontology $\mathcal{O}$ into a searchable vector index. The representations are encoded into dense numerical vectors using a frozen feature extractor $E(\cdot)$. Specifically, we deploy \texttt{Qwen3-Embedding-4B}~\cite{qwen3embedding}~\footnote{\url{https://huggingface.co/Qwen/Qwen3-Embedding-4B}} across all experimental setups to project all target ontological entities into a shared high-dimensional vector space: $\mathbf{v}_i = E(\text{text}(c_i)) \in \mathbb{R}^d$. The resulting embeddings are stored in a dense vector database indexed for fast nearest-neighbor lookups using cosine similarity.

\paragraph{Similarity-Based Retrieval.} During inference, an input candidate query $q$ (e.g., a candidate lexical term $L$ or a candidate pair of terms) is converted into an embedding using the same feature extractor $\mathbf{q} = E(\text{text}(q))$. Rather than relying on iterative multi-hop expansion or retriever-fine-tuning, we employ a single-pass dense retrieval strategy. The system computes the cosine similarity between the query vector $\mathbf{q}$ and all entity vectors $\mathbf{v}_i$ in the pre-indexed ontology base. We retrieve the top-$k$ nearest ontological entities $\mathcal{C}_q = \{c_1, c_2, \dots, c_k\}$ that yield the highest alignment scores $\text{Sim}(\mathbf{q}, \mathbf{v}_i) = \frac{\mathbf{q} \cdot \mathbf{v}_i}{\|\mathbf{q}\|_2 \|\mathbf{v}_i\|_2}$. For all experiments in this study, we fix $k=10$. The textual metadata associated with these top-$k$ entities forms the ground-truth reference context injected into the downstream prompt.

\paragraph{Task-Specific LLM Generation.} The final phase constructs a structured, task-specific instruction prompt combining three distinct elements: (1) a system role specification defining the target OL task rules, (2) the retrieved top-$k$ background ontological entities $\mathcal{C}_q$ serving as candidate grounded choices, and (3) the specific target input (e.g., lexical term $L$ for Term Typing or term pair $(L_a, L_b)$ for Taxonomy Discovery). The complete prompt is passed to the target LLM under zero-shot conditions with identical decoding hyperparameters across model sizes. To guarantee reliable evaluation, the model is constrained to generate structured output conforming to the expected formal schema for each task:
\begin{itemize}
    \item \textit{Term Typing:} The LLM selects the single most accurate ontological class $T$ from the candidate set $\mathcal{C}_q$ that encapsulates the lexical term $L$.
    \item \textit{Taxonomy Discovery:} The LLM determines whether a direct subsumption relationship (\texttt{is-a}) exists between the concepts associated with $L_a$ and $L_b$ within the retrieved context.
    \item \textit{Non-Taxonomic Relationship Extraction:} Given head term $h$, candidate relation $r$, and tail term $t$, the LLM predicts whether the relation holds based on the provided domain context.
\end{itemize}
By keeping the embedding model, vector index, top-$k$ parameter ($k=10$), and prompt templates strictly constant across all runs, any observed variation in task performance directly reflects the impact of model parameter scale and underlying architecture.

\section{Experimental Setup}



\begin{table}[t]
\centering
\caption{LLMs evaluated in this work from the Qwen3.5, Qwen3.6, and GPT families.}
\label{tab:llms}
\small

\resizebox{\textwidth}{!}{%
\begin{tabular}{l l l l l l}
\hline
\textbf{Model} &
\textbf{Family} &
\textbf{Architecture} &
\textbf{Scale} &
\textbf{Availability} &
\textbf{Model Page} \\
\hline

Qwen3.5-0.8B &
Qwen3.5 &
Dense &
0.8B &
Open &
\url{https://huggingface.co/Qwen/Qwen3.5-0.8B} \\

Qwen3.5-2B &
Qwen3.5 &
Dense &
2B &
Open &
\url{https://huggingface.co/Qwen/Qwen3.5-2B} \\

Qwen3.5-4B &
Qwen3.5 &
Dense &
4B &
Open &
\url{https://huggingface.co/Qwen/Qwen3.5-4B} \\

Qwen3.5-9B &
Qwen3.5 &
Dense &
9B &
Open &
\url{https://huggingface.co/Qwen/Qwen3.5-9B} \\

Qwen3.5-27B &
Qwen3.5 &
Dense &
27B &
Open &
\url{https://huggingface.co/Qwen/Qwen3.5-27B} \\

Qwen3.5-35B-A3B &
Qwen3.5 &
MoE &
35B (3B active) &
Open &
\url{https://huggingface.co/Qwen/Qwen3.5-35B-A3B} \\

Qwen3.5-122B-A10B &
Qwen3.5 &
MoE &
122B (10B active) &
Open &
\url{https://huggingface.co/Qwen/Qwen3.5-122B-A10B} \\

\hline

Qwen3.6-27B &
Qwen3.6 &
Dense &
27B &
Open &
\url{https://huggingface.co/Qwen/Qwen3.6-27B} \\

Qwen3.6-35B-A3B &
Qwen3.6 &
MoE &
35B (3B active) &
Open &
\url{https://huggingface.co/Qwen/Qwen3.6-35B-A3B} \\

\hline

GPT-5.5 &
GPT &
Proprietary &
-- &
Closed &
\url{https://developers.openai.com/api/docs/models/gpt-5.5} \\

GPT-5.6-Luna &
GPT &
Proprietary &
-- &
Closed &
\url{https://developers.openai.com/api/docs/models/gpt-5.6-luna} \\

GPT-5.6-Terra &
GPT &
Proprietary &
-- &
Closed &
\url{https://developers.openai.com/api/docs/models/gpt-5.6-terra} \\

GPT-5.6-Sol &
GPT &
Proprietary &
-- &
Closed &
\url{https://developers.openai.com/api/docs/models/gpt-5.6-sol} \\

\hline
\end{tabular}%
}
\end{table}

\begin{table}[t]
\caption{Selected ontological datasets ground truth statistics. The \textit{Terms No.} is the number of terms extracted from the given ontology, whereas the \textit{Types No.} is the number of classes, and \textit{Relations No.} is the number of non-is-a relations.}
\label{tab:datasets}
\centering
\resizebox{\textwidth}{!}{%
    \begin{tabular}{l|l|r|r|r|r|r|r}
    \hline
    \textbf{Domain} & \textbf{Ontology} & \textbf{Terms No.} & \textbf{Types No.} & \textbf{Relations No.} & \textbf{Term Typing} & \textbf{Taxonomy Discovery} & \textbf{Non-Taxonomic RE} \\
    \hline
     Material Science and Engineering & MatWerk~\cite{beygi2025nfdi} & 29 & 313 & 2 & 29 & 369 & 12 \\
      & MDS~\cite{rajamohan2025materials} & - & 297 & 3 & - & 351 & 128 \\
    \hline
     Medical & OBI~\cite{OBI} & 53 & 4,934 & 3 & 53 & 2,368 & - \\
      & MFOEM~\cite{hastings2011dispositions,hastings2012annotating} & 19 & 621 & 2 & 19 & 837 & 20 \\
    \hline
\end{tabular}
}
\end{table}
\paragraph{Large Language Models.} We evaluate a diverse set of open- and closed-source LLMs spanning multiple parameter scales and architectural designs. The benchmark includes dense and Mixture-of-Experts (MoE) variants from the Qwen3.5 and Qwen3.6 families, as well as frontier OpenAI GPT models, enabling a systematic investigation of how model size and architecture influence OL performance. \autoref{tab:llms} summarizes the evaluated models and their key characteristics, where all are supported by OntoLearner for reproducible experimentation and benchmarking.

\paragraph{Datasets.} For experimental purposes, we select four established ontologies from two scientific domains: materials science and engineering and the biomedical domain. These domains are selected to evaluate LLM-scale effects across heterogeneous scientific areas with different conceptual structures and levels of complexity. The selected ontologies include MatWerk~\cite{beygi2025nfdi} and MDS~\cite{rajamohan2025materials} from the materials science domain, and OBI~\cite{OBI} and MFOEM~\cite{hastings2011dispositions,hastings2012annotating} from the biomedical domain.  For each ontology, we used OntoLearner benchmark collections hosted in Hugging Face~\footnote{OntoLearner Benchmark Collections: \url{https://huggingface.co/collections/SciKnowOrg/ontolearner-benchmarking}} to construct evaluation tasks covering term typing, taxonomy discovery, and non-taxonomic relationship extraction. \autoref{tab:datasets} summarizes the ground truth statistics of the selected datasets. Moreover, to manage experiments, we used the OntoLearner \texttt{train\_test\_split} strategy to preserve only 20\% of the ontological data for OBI experiments.

\section{Results}

\begin{table}[t]
    \caption{Results on the medical ontologies.}
    \label{tab:medical_results}
    \centering
    \resizebox{\textwidth}{!}{ 
    \begin{tabular}{l c c c | c c c | c c c | c c c | c c c }
        \hline
        \multirow{3}{*}{\textbf{LLMs}} & \multicolumn{9}{c|}{\textbf{MFOEM}} & \multicolumn{6}{c}{\textbf{OBI}} \\
         \cline{2-16}
         & \multicolumn{3}{c|}{\textbf{Term Typing}} & \multicolumn{3}{c|}{\textbf{Taxonomy Discovery}} & \multicolumn{3}{c|}{\textbf{Non-Taxonomic RE}} & \multicolumn{3}{c|}{\textbf{Term Typing}} & \multicolumn{3}{c}{\textbf{Taxonomy Discovery}}  \\
         \cline{2-16}
         & \textbf{Prec} & \textbf{Rec} & \textbf{F1} & \textbf{Prec} & \textbf{Rec} & \textbf{F1} & \textbf{Prec} & \textbf{Rec} & \textbf{F1} & \textbf{Prec} & \textbf{Rec} & \textbf{F1} & \textbf{Prec} & \textbf{Rec} & \textbf{F1} \\
         \hline
        Qwen3.5-0.8B & 25.00 & 100.00 & 40.00 & 4.84 & 61.82 & 8.98 & 3.91 & 75.00 & 7.43 & 9.06 & 90.57 & 16.47 & 3.29 & 52.24 & 6.19   \\
        Qwen3.5-2B & 25.00 & 100.00 & 40.00 & 4.84 & 61.82 & 8.98 & 3.89 & 75.00 & 7.39 & 9.06 & 90.57 & 16.47 & 3.29 & 52.24 & 6.19   \\
        Qwen3.5-4B & 25.33 & 100.00 & 40.43 & 4.84 & 61.82 & 8.98 & 3.76 & 70.00 & 7.14 & 9.28 & 90.57 & 16.84 & 3.29 & 52.24 & 6.19  \\
        Qwen3.5-9B & 27.27 & 94.74 & 42.35 & 4.93 & 61.82 & 9.12 & 5.19 & 75.00 & 9.71 & 11.03 & 90.57 & 19.67 & 3.53 & 52.24 & 6.61  \\
        Qwen3.5-27B & 41.67 & 52.63 & 46.51 & 8.69 & 60.61 & 15.20 & 6.07 & 75.00 & 11.24 & 18.75 & 90.57 & 31.07 & 6.53 & 51.79 & 11.59   \\
        Qwen3.5-35B-A3B & 28.33 & 89.47 & 43.04 & 11.38 & 59.85 & 19.12 & 3.97 & 75.00 & 7.54 & 18.68 & 90.57 & 30.97 & 10.41 & 50.78 & 17.28  \\
        Qwen3.5-122B-A10B & 26.56 & 89.47 & 40.96 & 15.38 & 59.24 & 24.41 & 4.23 & 75.00 & 8.00 & 16.44 & 90.57 & 27.83 & 14.16 & 50.03 & 22.07   \\
        \hline
        Qwen3.6-27B & 35.56 & 84.21 & 50.00 & 11.07 & 60.61 & 18.73 & 4.41 & 75.00 & 8.33 & 13.30 & 90.57 & 23.19 & 9.92 & 51.54 & 16.63   \\
        Qwen3.6-35B-A3B & 26.76 & 100.00 & 42.22 & 9.16 & 59.09 & 15.86 & 3.95 & 75.00 & 7.50 & 9.56 & 90.57 & 17.30 & 7.58 & 50.88 & 13.20   \\
        \hline
        GPT-5.5 & 70.00 & 73.68 & \textbf{71.79} & 33.89 & 55.30 & \textbf{42.03} & 9.52 & 70.00 & \textbf{16.77} & 70.31 & 84.91 & 76.92 & 28.90 & 48.01 & \textbf{36.08}  \\
        GPT-5.6-Luna & 100.00 & 5.26 & 10.00 & 35.22 & 48.94 & 40.96 & 10.13 & 40.00 & 16.16 & 52.70 & 73.58 & 61.42 & 27.91 & 44.53 & 34.32  \\
        GPT-5.6-Terra & 71.43 & 26.32 & 38.46 & 31.41 & 53.48 & 39.57 & 8.48 & 70.00 & 15.14 & 51.11 & 86.79 & 64.34 & 30.06 & 45.09 & 36.07   \\
        GPT-5.6-Sol & 58.33 & 36.84 & 45.16 & 27.47 & 55.61 & 36.77 & 8.22 & 60.00 & 14.46 & 71.21 & 88.68 & \textbf{78.99} & 26.41 & 48.31 & 34.15   \\
        \hline
    \end{tabular}
    }
\end{table}

\begin{table}[]
    \caption{Results on the material science and engineering ontologies.}
    \label{tab:materials_results}
    \centering
    \resizebox{\textwidth}{!}{ 
    \begin{tabular}{l c c c | c c c | c c c | c c c | c c c }
        \hline
        \multirow{3}{*}{\textbf{LLMs}} & \multicolumn{9}{c|}{\textbf{MatWerk}} & \multicolumn{6}{c}{\textbf{MDS}} \\
         \cline{2-16}
         & \multicolumn{3}{c|}{\textbf{Term Typing}} & \multicolumn{3}{c|}{\textbf{Taxonomy Discovery}} & \multicolumn{3}{c|}{\textbf{Non-Taxonomic RE}} &  \multicolumn{3}{c|}{\textbf{Taxonomy Discovery}} & \multicolumn{3}{c}{\textbf{Non-Taxonomic RE}} \\
         \cline{2-16}
         & \textbf{Prec} & \textbf{Rec} & \textbf{F1} & \textbf{Prec} & \textbf{Rec} & \textbf{F1} & \textbf{Prec} & \textbf{Rec} & \textbf{F1} & \textbf{Prec} & \textbf{Rec} & \textbf{F1} & \textbf{Prec} & \textbf{Rec} & \textbf{F1} \\
         \hline
        Qwen3.5-0.8B & 14.29 & 100.00 & 25.00 & 4.25 & 60.66 & 7.94 & 3.60 & 81.82 & 6.90 &   1.36 & 18.95 & 2.54 & 0.62 & 27.34 & 1.22 \\
        Qwen3.5-2B & 14.29 & 100.00 & 25.00 & 4.22 & 60.33 & 7.89 & 3.60 & 81.82 & 6.90 &   1.39 & 19.28 & 2.59 & 0.61 & 26.56 & 1.19 \\
        Qwen3.5-4B & 14.65 & 100.00 & 25.55 & 4.22 & 60.33 & 7.89 & 3.35 & 72.73 & 6.40 &   1.39 & 19.28 & 2.59 & 0.62 & 26.56 & 1.20 \\
        Qwen3.5-9B & 22.22 & 96.55 & 36.13 & 4.52 & 60.33 & 8.40 & 5.03 & 81.82 & 9.47 &  1.42 & 19.28 & 2.64 & 0.51 & 21.09 & 0.99 \\
        Qwen3.5-27B & 43.86 & 86.21 & 58.14 & 6.76 & 59.34 & 12.13 & 6.62 & 81.82 & 12.24 &   3.24 & 18.63 & 5.52 & 0.56 & 14.06 & 1.07 \\
        Qwen3.5-35B-A3B & 20.28 & 100.00 & 33.72 & 11.34 & 58.03 & 18.97 & 3.63 & 81.82 & 6.95 &  2.77 & 16.99 & 4.77 & 0.61 & 26.56 & 1.20 \\
        Qwen3.5-122B-A10B & 24.78 & 96.55 & 39.44 & 14.50 & 56.07 & 23.05 & 3.90 & 81.82 & 7.44 &  5.65 & 13.73 & 8.01 & 0.65 & 26.56 & 1.27 \\
        \hline
        Qwen3.6-27B & 23.01 & 89.66 & 36.62 & 9.48 & 60.33 & 16.38 & 4.15 & 81.82 & 7.89 & 3.60 & 16.99 & 5.95 & 0.61 & 26.56 & 1.19 \\
        Qwen3.6-35B-A3B & 15.76 & 100.00 & 27.23 & 6.16 & 58.03 & 11.14 & 3.63 & 81.82 & 6.95 &   2.79 & 15.69 & 4.74 & 0.63 & 26.56 & 1.23 \\
        \hline
        GPT-5.5 & 70.59 & 82.76 & \textbf{76.19} & 29.70 & 48.20 & 36.75 & 8.25 & 72.73 & 14.81 &  16.88 & 12.75 & \textbf{14.53} & 2.80 & 7.81 & \textbf{4.12} \\
        GPT-5.6-Luna & 73.68 & 48.28 & 58.33 & 33.88 & 47.21 & \textbf{39.45} & 13.33 & 36.36 & \textbf{19.51} &  20.56 & 7.19 & 10.65 & 0.00 & 0.00 & 0.00 \\
        GPT-5.6-Terra & 69.23 & 62.07 & 65.45 & 30.95 & 43.93 & 36.31 & 7.63 & 81.82 & 13.95 &   18.64 & 7.19 & 10.38 & 1.08 & 7.81 & 1.90 \\
        GPT-5.6-Sol & 76.92 & 68.97 & 72.73 & 24.23 & 51.80 & 33.02 & 7.92 & 72.73 & 14.29 &   13.17 & 12.09 & 12.61 & 1.08 & 7.03 & 1.87 \\
        \hline
    \end{tabular}
    }
\end{table}

The experimental results for medical ontologies are presented in \autoref{tab:medical_results}, and for material science and engineering domain ontologies are represented in \autoref{tab:materials_results} for three core LLMs4OL paradigm tasks using precision, recall, and F1-scores.  Overall, the empirical evaluation reveals three insights: (1) increasing parameter scale in open-weight models primarily enhances the decision boundary rather than recall; (2) dense models at medium scale (e.g., 27B) often outperform much larger MoE models on the term typing task; and (3) domain complexity creates severe performance bottlenecks, particularly in the non-taxonomic relation extraction task on specialized domains.

\paragraph{Model Scale Dynamics in Open-Weight Families.} Evaluating the Qwen parameter progression (0.8B to 122B) demonstrates that scaling model capacity differs across tasks and parameter sizes.  More technically:
\begin{itemize}
    \item \underline{\textit{Precision Deficits in Small Models ($<9$B).}} Small-scale open-weight models (Qwen3.5-0.8B, 2B, and 4B) exhibit a persistent pattern in performance. On Term Typing across MFOEM, OBI, and MatWerk, these models achieve near-perfect or complete recall ($90.57\%\text{--}100.00\%$) but with a poor precision ($9.06\%\text{--}25.33\%$). In taxonomy discovery and non-taxonomic relation extraction, precision collapses further to $1.36\%\text{--}4.84\%$. This behavior indicates that the Qwen family of models with fewer than 9B parameters lacks the decision boundary calibration required to reject negative candidate classes provided within the RAG.

    \item \underline{\textit{The 27B Parameter Turning Point.}} The transition from 9B to 27B parameters marks the primary performance inflection point within the open-weight model family. On MatWerk Term Typing, Qwen3.5-27B achieves an F1-score of $58.14\%$, representing a $+22.01$ percentage point gain over Qwen3.5-9B ($36.13\%$). This improvement is driven by a doubling in precision (from $22.22\%$ to $43.86\%$) while maintaining high recall ($86.21\%$). A similar precision-driven gain is observed on OBI Term Typing ($31.07\%$ vs.\ $19.67\%$ F1).    

    \item \underline{\textit{Dense vs. Mixture-of-Experts Trade-offs.}} A comparison between dense and sparse MoE architectures reveals clear task-dependent trade-offs:
    \begin{itemize}
        \item \textit{Term Typing:} Dense models demonstrate superior active parameter efficiency. The dense Qwen3.5-27B consistently outperforms the larger Qwen3.5-122B-A10B MoE model on term typing across MFOEM ($46.51\%$ vs.\ $40.96\%$ F1) and MatWerk ($58.14\%$ vs.\ $39.44\%$ F1). High active parameter density appears essential for the term typing task.
        \item \textit{Taxonomy Discovery:} Conversely, MoE architectures excel at hierarchical relational reasoning. On MFOEM and MatWerk, Qwen3.5-122B-A10B achieves the highest F1-scores among all open-weight models ($24.41\%$ and $23.05\%$, respectively), outperforming its dense 27B counterpart ($15.20\%$ and $12.13\%$). The expanded total capacity of MoE models provides the enhanced capability necessary to evaluate subsumption relationships across paired concepts.
    \end{itemize}
    
    \item \underline{\textit{Version-Specific Alignment Anomaly (Qwen 3.5 vs. 3.6).}} Comparing release versions reveals an unexpected performance regression in Qwen3.6-27B relative to Qwen3.5-27B on MatWerk Term Typing ($36.62\%$ vs.\ $58.14\%$ F1). This drop suggests that architectural modifications or alignment fine-tuning introduced in version 3.6 altered the model's sensitivity to specialized engineering terminologies, reinforcing that parameter scale alone does not guarantee monotonic performance improvements across distinct model releases.
\end{itemize}

\paragraph{Frontier Model Performance and the Alignment Penalty.}Proprietary frontier models (GPT-5.5 and GPT-5.6 variants) maintain a substantial performance margin over open-weight baselines, achieving peak F1-scores of $71.79\%$ on MFOEM term typing and $76.19\%$ on MatWerk term typing. However, cross-model comparison within the GPT family highlights significant alignment-induced variations.
\begin{itemize}
    \item \underline{\textit{GPT-5.5 Superiority.}} Across nearly all evaluated tasks and domains, GPT-5.5 outperforms its newer GPT-5.6 successors (Luna, Terra, and Sol). It provides a well-calibrated trade-off between precision and recall (e.g., $70.00\%$ precision and $73.68\%$ recall on MFOEM Term Typing), demonstrating robust generalization across domain shifts.
    \item \underline{\textit{Over-Conservatism in GPT-5.6 Lineage.}} The GPT-5.6 family demonstrates an over-conservative bias, prioritizing precision over recall. This dynamic is most visible in \textbf{GPT-5.6-Luna}:
    \begin{itemize}
        \item On MFOEM term typing, GPT-5.6-Luna yields $100\%$ precision but collapses to $5.26\%$ recall, resulting in a low F1-score of $10\%$.
        \item On MDS Non-Taxonomic Relation Extraction, GPT-5.6-Luna fails entirely ($0\%$ Precision, Recall, and F1), declining to output positive relationship assertions.
    \end{itemize}
    This behavior suggests that the 5.6 release rendered certain model variants highly prone to false negatives when faced with domain-specific knowledge.
\end{itemize}

\begin{figure}[t]
    \centering
    \includegraphics[width=\linewidth]{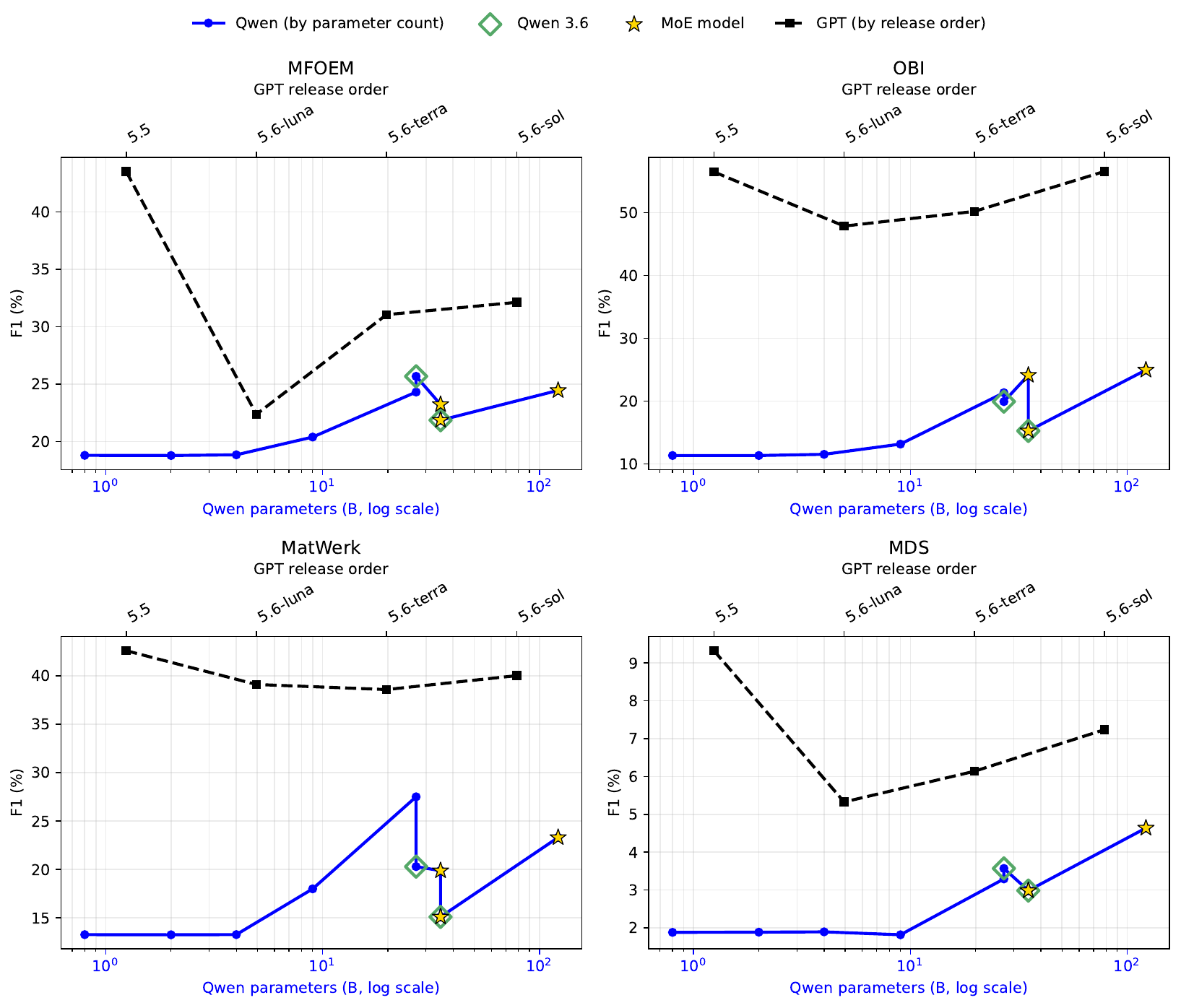}
    \caption{F1-score performance across model scale and release lineages for the evaluated biomedical (MFOEM, OBI) and materials science and engineering (MatWerk, MDS) ontologies. Qwen parameter scale is plotted on a logarithmic scale ($10^0$ to $10^2$ billion parameters), highlighting architectural transitions (dense vs. sparse MoE and Qwen 3.5 vs. 3.6), alongside performance trajectories across the GPT release sequence (GPT-5.5 to GPT-5.6 variants).}
    \label{fig:f1_vs_scale}
\end{figure}

\paragraph{Scale Sensitivity Across Task Complexity and Domains.} To study how model size affects performance scaling, we examine how task complexity and domain abstraction influence the benefits of increasing parameter capacity (shown by the scaling curves in \autoref{fig:f1_vs_scale}).
\begin{itemize}
    \item \underline{\textit{Task Complexity Changes the Required Scale.}} The parameter size needed to achieve useful performance increases with task structural complexity ($\text{Term Typing} \prec \text{Taxonomy Discovery} \prec \text{Non-Taxonomic RE}$):
    \begin{itemize}
        \item \textit{Low-Floor Task (Term Typing):} This task only requires selecting the correct candidate from a top-$k$ list. As a result, scaling provides strong improvements at smaller model sizes ($0.8\text{B} \rightarrow 27\text{B}$ achieves a $+33.14$ F1 gain on MatWerk).
        \item \textit{Moderate-Floor Task (Taxonomy Discovery):} This task requires reasoning over relationships between multiple concepts. Small models ($<9$B) reach a performance plateau below $9\%$ F1, while noticeable improvements appear only at $27\text{B}+$ and frontier-scale models ($36\%\text{--}42\%$ F1).
        \item \textit{High-Floor Task (Non-Taxonomic RE):} Detecting arbitrary semantic relations ($h, r, t$) is a highly complex task. Increasing parameters for open-weight models provides limited improvements, reaching only $12.24\%$ F1 for Qwen3.5-27B on MatWerk. Even frontier-scale models achieve only modest F1 scores ($16.77\%\text{--}19.51\%$), showing that increasing model size alone is not enough to solve multi-relational semantic spaces under zero-shot RAG settings.
    \end{itemize}
    \item \underline{\textit{Domain Abstraction Creates Scale Resistance in MDS.}} Comparing different domains shows that the level of abstraction affects how much model scaling can improve performance. Scaling consistently improves results in biomedical domains (MFOEM, OBI) and material synthesis (MatWerk). However, the highly abstract Materials Data Science (MDS) domain shows strong \textit{scale resistance}. For MDS Non-Taxonomic Relation Extraction, increasing Qwen model size from 0.8B to 122B provides almost no improvement ($1.22\%$ vs.\ $1.27\%$ F1), while frontier GPT models achieve only $4.12\%$ F1. Highly abstract and non-standardized conceptual schemas create difficulties that cannot be solved by parameter scaling alone.
\end{itemize}

\section{Conclusion}
In this work, we conducted a systematic study of how LLM scale influences OL performance across different tasks and scientific domains. Using a unified RAG-based evaluation framework, we benchmarked multiple model sizes from the Qwen and GPT families on LLMs4OL paradigm tasks. Our results show that increasing model size does not always lead to proportional improvements. Larger models generally improve precision and decision-making ability, but the benefits strongly depend on task complexity and domain characteristics. Term typing benefits most from scaling, while taxonomy discovery requires larger capacity for relational reasoning. In contrast, non-taxonomic relation extraction remains challenging even for frontier-scale models, particularly in highly abstract domains such as Materials Data Science. We also observe that architectural choices, such as dense versus MoE models, and model alignment strategies can significantly affect performance beyond parameter count alone. These findings highlight that model scale is only one factor determining OL capability. Effective LLM-based OL requires considering task structure, domain abstraction, and model architecture when selecting an appropriate model. Our OntoLearner infrastructure provides empirical guidance for choosing LLMs for ontology engineering and establishes a foundation for future research on scalable and reproducible OL systems.

\begin{acknowledgments}
This work is supported by the \href{https://www.nfdi4datascience.de/}{NFDI4DataScience initiative} (DFG, German Research Foundation, Grant ID: 460234259).

\end{acknowledgments}

\section*{Declaration on Generative AI}
In preparing this manuscript, generative AI tools, specifically ChatGPT, were used solely for grammar checking, spelling checks, and readability of some sentences.  All suggested changes were carefully reviewed and adapted by the authors to ensure accuracy and appropriateness. The scientific content, research design, analysis, and conclusions were developed and verified exclusively by the authors without AI involvement. The use of ChatGPT was limited to enhancing the presentation of the work.

\bibliography{sample-ceur}

\begin{thebibliography}{64}
\expandafter\ifx\csname natexlab\endcsname\relax\def\natexlab#1{#1}\fi
\providecommand{\url}[1]{\texttt{#1}}
\providecommand{\href}[2]{#2}
\providecommand{\path}[1]{#1}
\providecommand{\DOIprefix}{doi:}
\providecommand{\ArXivprefix}{arXiv:}
\providecommand{\URLprefix}{URL: }
\providecommand{\Pubmedprefix}{pmid:}
\providecommand{\doi}[1]{\href{http://dx.doi.org/#1}{\path{#1}}}
\providecommand{\Pubmed}[1]{\href{pmid:#1}{\path{#1}}}
\providecommand{\bibinfo}[2]{#2}
\ifx\xfnm\relax \def\xfnm[#1]{\unskip,\space#1}\fi
\bibitem[{Bock et~al.(2022)Bock, Datlinger, Chardon, Coelho, Dong, Lawson, Lu, Maroc, Norman, Song et~al.}]{bock2022high}
\bibinfo{author}{C.~Bock}, \bibinfo{author}{P.~Datlinger}, \bibinfo{author}{F.~Chardon}, \bibinfo{author}{M.~A. Coelho}, \bibinfo{author}{M.~B. Dong}, \bibinfo{author}{K.~A. Lawson}, \bibinfo{author}{T.~Lu}, \bibinfo{author}{L.~Maroc}, \bibinfo{author}{T.~M. Norman}, \bibinfo{author}{B.~Song}, et~al.,
\newblock \bibinfo{title}{High-content crispr screening},
\newblock \bibinfo{journal}{Nature Reviews Methods Primers} \bibinfo{volume}{2} (\bibinfo{year}{2022}) \bibinfo{pages}{8}.
\bibitem[{Tom et~al.(2024)Tom, Schmid, Baird, Cao, Darvish, Hao, Lo, Pablo-Garc{\'\i}a, Rajaonson, Skreta et~al.}]{tom2024self}
\bibinfo{author}{G.~Tom}, \bibinfo{author}{S.~P. Schmid}, \bibinfo{author}{S.~G. Baird}, \bibinfo{author}{Y.~Cao}, \bibinfo{author}{K.~Darvish}, \bibinfo{author}{H.~Hao}, \bibinfo{author}{S.~Lo}, \bibinfo{author}{S.~Pablo-Garc{\'\i}a}, \bibinfo{author}{E.~M. Rajaonson}, \bibinfo{author}{M.~Skreta}, et~al.,
\newblock \bibinfo{title}{Self-driving laboratories for chemistry and materials science},
\newblock \bibinfo{journal}{Chemical Reviews} \bibinfo{volume}{124} (\bibinfo{year}{2024}) \bibinfo{pages}{9633--9732}.
\bibitem[{Wilkinson et~al.(2016)Wilkinson, Dumontier, Aalbersberg, Appleton, Axton, Baak, Blomberg, Boiten, da~Silva~Santos, Bourne et~al.}]{wilkinson2016fair}
\bibinfo{author}{M.~D. Wilkinson}, \bibinfo{author}{M.~Dumontier}, \bibinfo{author}{I.~J. Aalbersberg}, \bibinfo{author}{G.~Appleton}, \bibinfo{author}{M.~Axton}, \bibinfo{author}{A.~Baak}, \bibinfo{author}{N.~Blomberg}, \bibinfo{author}{J.-W. Boiten}, \bibinfo{author}{L.~B. da~Silva~Santos}, \bibinfo{author}{P.~E. Bourne}, et~al.,
\newblock \bibinfo{title}{The fair guiding principles for scientific data management and stewardship},
\newblock \bibinfo{journal}{Scientific data} \bibinfo{volume}{3} (\bibinfo{year}{2016}) \bibinfo{pages}{1--9}.
\bibitem[{Auer et~al.(2025)Auer, D’Souza, Farfar, Jaradeh, Jiomekong, Karras, Oelen, Snyder, Stocker, and Vogt}]{auer2025open}
\bibinfo{author}{S.~Auer}, \bibinfo{author}{J.~D’Souza}, \bibinfo{author}{K.~E. Farfar}, \bibinfo{author}{M.~Y. Jaradeh}, \bibinfo{author}{A.~Jiomekong}, \bibinfo{author}{O.~Karras}, \bibinfo{author}{A.~Oelen}, \bibinfo{author}{L.~Snyder}, \bibinfo{author}{M.~Stocker}, \bibinfo{author}{L.~Vogt},
\newblock \bibinfo{title}{Open research knowledge graph: A large-scale neuro-symbolic knowledge organization system},
\newblock in: \bibinfo{booktitle}{Handbook on Neurosymbolic AI and Knowledge Graphs}, \bibinfo{publisher}{IOS Press}, \bibinfo{year}{2025}, pp. \bibinfo{pages}{385--420}.
\bibitem[{Wang et~al.(2023)Wang, Fu, Du, Gao, Huang, Liu, Chandak, Liu, Van~Katwyk, Deac et~al.}]{wang2023scientific}
\bibinfo{author}{H.~Wang}, \bibinfo{author}{T.~Fu}, \bibinfo{author}{Y.~Du}, \bibinfo{author}{W.~Gao}, \bibinfo{author}{K.~Huang}, \bibinfo{author}{Z.~Liu}, \bibinfo{author}{P.~Chandak}, \bibinfo{author}{S.~Liu}, \bibinfo{author}{P.~Van~Katwyk}, \bibinfo{author}{A.~Deac}, et~al.,
\newblock \bibinfo{title}{Scientific discovery in the age of artificial intelligence},
\newblock \bibinfo{journal}{Nature} \bibinfo{volume}{620} (\bibinfo{year}{2023}) \bibinfo{pages}{47--60}.
\bibitem[{Eger et~al.(2025)Eger, Cao, D'Souza, Geiger, Greisinger, Gross, Hou, Krenn, Lauscher, Li et~al.}]{eger2025transforming}
\bibinfo{author}{S.~Eger}, \bibinfo{author}{Y.~Cao}, \bibinfo{author}{J.~D'Souza}, \bibinfo{author}{A.~Geiger}, \bibinfo{author}{C.~Greisinger}, \bibinfo{author}{S.~Gross}, \bibinfo{author}{Y.~Hou}, \bibinfo{author}{B.~Krenn}, \bibinfo{author}{A.~Lauscher}, \bibinfo{author}{Y.~Li}, et~al.,
\newblock \bibinfo{title}{Transforming science with large language models: A survey on ai-assisted scientific discovery, experimentation, content generation, and evaluation},
\newblock \bibinfo{journal}{arXiv preprint arXiv:2502.05151}  (\bibinfo{year}{2025}).
\bibitem[{Kendall and McGuinness(2019)}]{kendall2019ontology}
\bibinfo{author}{E.~F. Kendall}, \bibinfo{author}{D.~L. McGuinness}, \bibinfo{title}{Ontology engineering}, \bibinfo{publisher}{Morgan \& Claypool Publishers}, \bibinfo{year}{2019}.
\bibitem[{Giglou et~al.(2023)Giglou, D’Souza, and Auer}]{giglou2023llms4ol}
\bibinfo{author}{H.~B. Giglou}, \bibinfo{author}{J.~D’Souza}, \bibinfo{author}{S.~Auer},
\newblock \bibinfo{title}{Llms4ol: Large language models for ontology learning},
\newblock \bibinfo{journal}{Lecture Notes in Computer Science}  (\bibinfo{year}{2023}) \bibinfo{pages}{408--427}.
\bibitem[{Devlin et~al.(2019)Devlin, Chang, Lee, and Toutanova}]{bert}
\bibinfo{author}{J.~Devlin}, \bibinfo{author}{M.-W. Chang}, \bibinfo{author}{K.~Lee}, \bibinfo{author}{K.~Toutanova}, \bibinfo{title}{Bert: Pre-training of deep bidirectional transformers for language understanding}, \bibinfo{year}{2019}. \href{http://arxiv.org/abs/1810.04805}{{\tt arXiv:1810.04805}}.
\bibitem[{Lewis et~al.(2019)Lewis, Liu, Goyal, Ghazvininejad, Mohamed, Levy, Stoyanov, and Zettlemoyer}]{bart}
\bibinfo{author}{M.~Lewis}, \bibinfo{author}{Y.~Liu}, \bibinfo{author}{N.~Goyal}, \bibinfo{author}{M.~Ghazvininejad}, \bibinfo{author}{A.~Mohamed}, \bibinfo{author}{O.~Levy}, \bibinfo{author}{V.~Stoyanov}, \bibinfo{author}{L.~Zettlemoyer}, \bibinfo{title}{Bart: Denoising sequence-to-sequence pre-training for natural language generation, translation, and comprehension}, \bibinfo{year}{2019}. \href{http://arxiv.org/abs/1910.13461}{{\tt arXiv:1910.13461}}.
\bibitem[{Longpre et~al.(2023)Longpre, Hou, Vu, Webson, Chung, Tay, Zhou, Le, Zoph, Wei et~al.}]{longpre2023flan}
\bibinfo{author}{S.~Longpre}, \bibinfo{author}{L.~Hou}, \bibinfo{author}{T.~Vu}, \bibinfo{author}{A.~Webson}, \bibinfo{author}{H.~W. Chung}, \bibinfo{author}{Y.~Tay}, \bibinfo{author}{D.~Zhou}, \bibinfo{author}{Q.~V. Le}, \bibinfo{author}{B.~Zoph}, \bibinfo{author}{J.~Wei}, et~al.,
\newblock \bibinfo{title}{The flan collection: Designing data and methods for effective instruction tuning},
\newblock \bibinfo{journal}{arXiv preprint arXiv:2301.13688}  (\bibinfo{year}{2023}).
\bibitem[{Workshop et~al.(2023)Workshop, :, Scao, Fan, and et~al.}]{BLOOM}
\bibinfo{author}{B.~Workshop}, \bibinfo{author}{:}, \bibinfo{author}{T.~L. Scao}, \bibinfo{author}{A.~Fan}, \bibinfo{author}{et~al.}, \bibinfo{title}{Bloom: A 176b-parameter open-access multilingual language model}, \bibinfo{year}{2023}. \URLprefix \url{https://arxiv.org/abs/2211.05100}. \href{http://arxiv.org/abs/2211.05100}{{\tt arXiv:2211.05100}}.
\bibitem[{Touvron et~al.(2023)Touvron, Lavril, Izacard, Martinet, Lachaux, Lacroix, Rozi{\`e}re, Goyal, Hambro, Azhar et~al.}]{llama}
\bibinfo{author}{H.~Touvron}, \bibinfo{author}{T.~Lavril}, \bibinfo{author}{G.~Izacard}, \bibinfo{author}{X.~Martinet}, \bibinfo{author}{M.-A. Lachaux}, \bibinfo{author}{T.~Lacroix}, \bibinfo{author}{B.~Rozi{\`e}re}, \bibinfo{author}{N.~Goyal}, \bibinfo{author}{E.~Hambro}, \bibinfo{author}{F.~Azhar}, et~al.,
\newblock \bibinfo{title}{Llama: Open and efficient foundation language models},
\newblock \bibinfo{journal}{arXiv preprint arXiv:2302.13971}  (\bibinfo{year}{2023}).
\bibitem[{Dale(2021)}]{dale2021gpt}
\bibinfo{author}{R.~Dale},
\newblock \bibinfo{title}{Gpt-3: What’s it good for?},
\newblock \bibinfo{journal}{Natural Language Engineering} \bibinfo{volume}{27} (\bibinfo{year}{2021}) \bibinfo{pages}{113--118}.
\bibitem[{Gu et~al.(2021)Gu, Tinn, Cheng, Lucas, Usuyama, Liu, Naumann, Gao, and Poon}]{gu2021domain}
\bibinfo{author}{Y.~Gu}, \bibinfo{author}{R.~Tinn}, \bibinfo{author}{H.~Cheng}, \bibinfo{author}{M.~Lucas}, \bibinfo{author}{N.~Usuyama}, \bibinfo{author}{X.~Liu}, \bibinfo{author}{T.~Naumann}, \bibinfo{author}{J.~Gao}, \bibinfo{author}{H.~Poon},
\newblock \bibinfo{title}{Domain-specific language model pretraining for biomedical natural language processing},
\newblock \bibinfo{journal}{ACM Transactions on Computing for Healthcare (HEALTH)} \bibinfo{volume}{3} (\bibinfo{year}{2021}) \bibinfo{pages}{1--23}.
\bibitem[{Babaei~Giglou et~al.(2024)Babaei~Giglou, D’Souza, and Auer}]{BabaeiGiglou2024LLMs4OL}
\bibinfo{author}{H.~Babaei~Giglou}, \bibinfo{author}{J.~D’Souza}, \bibinfo{author}{S.~Auer},
\newblock \bibinfo{title}{Llms4ol 2024 overview: The 1st large language models for ontology learning challenge},
\newblock \bibinfo{journal}{Open Conference Proceedings} \bibinfo{volume}{4} (\bibinfo{year}{2024}) \bibinfo{pages}{3–16}. \DOIprefix\doi{10.52825/ocp.v4i.2473}.
\bibitem[{Babaei~Giglou et~al.(2025)Babaei~Giglou, D'Souza, Mihindukulasooriya, and Auer}]{BabaeiGiglou2025LLMs4OL}
\bibinfo{author}{H.~Babaei~Giglou}, \bibinfo{author}{J.~D'Souza}, \bibinfo{author}{N.~Mihindukulasooriya}, \bibinfo{author}{S.~Auer},
\newblock \bibinfo{title}{Llms4ol 2025 overview: The 2nd large language models for ontology learning challenge},
\newblock \bibinfo{journal}{Open Conference Proceedings} \bibinfo{volume}{6} (\bibinfo{year}{2025}). \DOIprefix\doi{10.52825/ocp.v6i.2913}.
\bibitem[{Peng et~al.(2024)Peng, Mou, Zhu, Sowe, and Decker}]{RWTHDBIS2024}
\bibinfo{author}{Y.~Peng}, \bibinfo{author}{Y.~Mou}, \bibinfo{author}{B.~Zhu}, \bibinfo{author}{S.~Sowe}, \bibinfo{author}{S.~Decker},
\newblock \bibinfo{title}{Rwth-dbis at llms4ol 2024 tasks a and b},
\newblock \bibinfo{journal}{Open Conference Proceedings} \bibinfo{volume}{4} (\bibinfo{year}{2024}).
\bibitem[{Hashemi et~al.(2024)Hashemi, {Karimi Manesh}, and Shamsfard}]{SKHNLP2024}
\bibinfo{author}{S.~Hashemi}, \bibinfo{author}{M.~{Karimi Manesh}}, \bibinfo{author}{M.~Shamsfard},
\newblock \bibinfo{title}{Skh-nlp at llms4ol 2024 task b: Taxonomy discovery in ontologies using bert and llama 3},
\newblock \bibinfo{journal}{Open Conference Proceedings} \bibinfo{volume}{4} (\bibinfo{year}{2024}).
\bibitem[{Phuttaamart et~al.(2024)Phuttaamart, Kertkeidkachorn, and Trongratsameethong}]{TheGhost2024}
\bibinfo{author}{T.~Phuttaamart}, \bibinfo{author}{N.~Kertkeidkachorn}, \bibinfo{author}{A.~Trongratsameethong},
\newblock \bibinfo{title}{The ghost at llms4ol 2024 task a: Prompt-tuning-based large language models for term typing},
\newblock \bibinfo{journal}{Open Conference Proceedings} \bibinfo{volume}{4} (\bibinfo{year}{2024}).
\bibitem[{Sanaei et~al.(2024)Sanaei, Azizi, and Babaei~Giglou}]{Phoenixes2024}
\bibinfo{author}{M.~Sanaei}, \bibinfo{author}{F.~Azizi}, \bibinfo{author}{H.~Babaei~Giglou},
\newblock \bibinfo{title}{Phoenixes at llms4ol 2024 tasks a, b, and c: Retrieval augmented generation for ontology learning},
\newblock \bibinfo{journal}{Open Conference Proceedings} \bibinfo{volume}{4} (\bibinfo{year}{2024}).
\bibitem[{Ymele and Jiomekong(2024)}]{TSOTSALearning2024}
\bibinfo{author}{C.~Ymele}, \bibinfo{author}{A.~Jiomekong},
\newblock \bibinfo{title}{Combining rules to large language model for ontology learning},
\newblock \bibinfo{journal}{Open Conference Proceedings} \bibinfo{volume}{4} (\bibinfo{year}{2024}).
\bibitem[{{Kumar Goyal} et~al.(2024){Kumar Goyal}, Singh, and {Shanker Tiwari}}]{SilpNLP2024}
\bibinfo{author}{P.~{Kumar Goyal}}, \bibinfo{author}{S.~Singh}, \bibinfo{author}{U.~{Shanker Tiwari}},
\newblock \bibinfo{title}{silp\_nlp at llms4ol 2024 tasks a, b, and c: Ontology learning through prompts with llms},
\newblock \bibinfo{journal}{Open Conference Proceedings} \bibinfo{volume}{4} (\bibinfo{year}{2024}).
\bibitem[{Rahnamoun and Shamsfard(2025)}]{SBUNLP2025}
\bibinfo{author}{R.~Rahnamoun}, \bibinfo{author}{M.~Shamsfard},
\newblock \bibinfo{title}{Sbu-nlp at llms4ol 2025 tasks a, b, and c: Stage-wise ontology construction through llms without any training procedure},
\newblock \bibinfo{journal}{Open Conference Proceedings}  (\bibinfo{year}{2025}).
\bibitem[{Beliaeva and Rahmatullaev(2025)}]{Alexbek2025}
\bibinfo{author}{A.~Beliaeva}, \bibinfo{author}{T.~Rahmatullaev},
\newblock \bibinfo{title}{Alexbek at llms4ol 2025 tasks a, b, and c: Heterogeneous llm methods for ontology learning (few-shot prompting, ensemble typing, and attention-based taxonomies)},
\newblock \bibinfo{journal}{Open Conference Proceedings}  (\bibinfo{year}{2025}).
\bibitem[{Goyal et~al.(2025)Goyal, Singh, and Tiwary}]{SilpNLP2025}
\bibinfo{author}{P.~Goyal}, \bibinfo{author}{S.~Singh}, \bibinfo{author}{U.~S. Tiwary},
\newblock \bibinfo{title}{silp\_nlp at llms4ol 2025 tasks a, b, c, and d: Clustering-based ontology learning using llms},
\newblock \bibinfo{journal}{Open Conference Proceedings}  (\bibinfo{year}{2025}).
\bibitem[{Latipov et~al.(2025)Latipov, Holenderski, and Meratnia}]{IRIS2025}
\bibinfo{author}{I.-A. Latipov}, \bibinfo{author}{M.~Holenderski}, \bibinfo{author}{N.~Meratnia},
\newblock \bibinfo{title}{Iris at llms4ol 2025 tasks b, c, and d: Enhancing ontology learning through data enrichment and type filtering},
\newblock \bibinfo{journal}{Open Conference Proceedings}  (\bibinfo{year}{2025}).
\bibitem[{Wiangnak et~al.(2025)Wiangnak, Prabhong, Phuttaamart, Kertkeidkachorn, and Shirai}]{DREAMLLMs2025}
\bibinfo{author}{P.~Wiangnak}, \bibinfo{author}{T.~Prabhong}, \bibinfo{author}{T.~Phuttaamart}, \bibinfo{author}{N.~Kertkeidkachorn}, \bibinfo{author}{K.~Shirai},
\newblock \bibinfo{title}{The dream-llms at llms4ol 2025 task b: A deliberation-based reasoning ensemble approach with multiple large language models for term typing in low-resource domains},
\newblock \bibinfo{journal}{Open Conference Proceedings}  (\bibinfo{year}{2025}).
\bibitem[{Fridouni and Sanaei(2025)}]{Phoenixes2025}
\bibinfo{author}{A.~E. Fridouni}, \bibinfo{author}{M.~Sanaei},
\newblock \bibinfo{title}{Phoenixes at llms4ol 2025 task a: Ontology learning with large language models reasoning},
\newblock \bibinfo{journal}{Open Conference Proceedings}  (\bibinfo{year}{2025}).
\bibitem[{Roche et~al.(2025)Roche, Gray, Murdock, and Crowder}]{ELLMO2025}
\bibinfo{author}{R.~Roche}, \bibinfo{author}{K.~Gray}, \bibinfo{author}{J.~Murdock}, \bibinfo{author}{D.~C. Crowder},
\newblock \bibinfo{title}{Ellmo at llms4ol 2025 tasks a and d: Llm-based term, type, and relationship extraction},
\newblock \bibinfo{journal}{Open Conference Proceedings}  (\bibinfo{year}{2025}).
\bibitem[{Kaplan et~al.(2020)Kaplan, McCandlish, Henighan, Brown, Chess, Child, Gray, Radford, Wu, and Amodei}]{kaplan2020scaling}
\bibinfo{author}{J.~Kaplan}, \bibinfo{author}{S.~McCandlish}, \bibinfo{author}{T.~Henighan}, \bibinfo{author}{T.~B. Brown}, \bibinfo{author}{B.~Chess}, \bibinfo{author}{R.~Child}, \bibinfo{author}{S.~Gray}, \bibinfo{author}{A.~Radford}, \bibinfo{author}{J.~Wu}, \bibinfo{author}{D.~Amodei},
\newblock \bibinfo{title}{Scaling laws for neural language models},
\newblock \bibinfo{journal}{arXiv preprint arXiv:2001.08361}  (\bibinfo{year}{2020}).
\bibitem[{Hoffmann et~al.(2022)Hoffmann, Borgeaud, Mensch, Buchatskaya, Cai, Rutherford, Casas, Hendricks, Welbl, Clark et~al.}]{hoffmann2022training}
\bibinfo{author}{J.~Hoffmann}, \bibinfo{author}{S.~Borgeaud}, \bibinfo{author}{A.~Mensch}, \bibinfo{author}{E.~Buchatskaya}, \bibinfo{author}{T.~Cai}, \bibinfo{author}{E.~Rutherford}, \bibinfo{author}{D.~d.~L. Casas}, \bibinfo{author}{L.~A. Hendricks}, \bibinfo{author}{J.~Welbl}, \bibinfo{author}{A.~Clark}, et~al.,
\newblock \bibinfo{title}{Training compute-optimal large language models},
\newblock \bibinfo{journal}{arXiv preprint arXiv:2203.15556}  (\bibinfo{year}{2022}).
\bibitem[{Wei et~al.(2022)Wei, Tay, Bommasani, Raffel, Zoph, Borgeaud, Yogatama, Bosma, Zhou, Metzler et~al.}]{wei2022emergent}
\bibinfo{author}{J.~Wei}, \bibinfo{author}{Y.~Tay}, \bibinfo{author}{R.~Bommasani}, \bibinfo{author}{C.~Raffel}, \bibinfo{author}{B.~Zoph}, \bibinfo{author}{S.~Borgeaud}, \bibinfo{author}{D.~Yogatama}, \bibinfo{author}{M.~Bosma}, \bibinfo{author}{D.~Zhou}, \bibinfo{author}{D.~Metzler}, et~al.,
\newblock \bibinfo{title}{Emergent abilities of large language models},
\newblock \bibinfo{journal}{arXiv preprint arXiv:2206.07682}  (\bibinfo{year}{2022}).
\bibitem[{Schaeffer et~al.(2023)Schaeffer, Miranda, and Koyejo}]{schaeffer2023mirage}
\bibinfo{author}{R.~Schaeffer}, \bibinfo{author}{B.~Miranda}, \bibinfo{author}{S.~Koyejo},
\newblock \bibinfo{title}{Are emergent abilities of llms a mirage?},
\newblock \bibinfo{journal}{Proc. NeurIPS} \bibinfo{volume}{23} (\bibinfo{year}{2023}).
\bibitem[{McKenzie et~al.(2023)McKenzie, Lyzhov, Pieler, Parrish, Mueller, Prabhu, McLean, Shen, Cavanagh, Gritsevskiy, Kauffman, Kirtland, Zhou, Zhang, Huang, Wurgaft, Weiss, Ross, Recchia, Liu, Liu, Tseng, Korbak, Kim, Bowman, and Perez}]{mckenzie2023inverse}
\bibinfo{author}{I.~R. McKenzie}, \bibinfo{author}{A.~Lyzhov}, \bibinfo{author}{M.~M. Pieler}, \bibinfo{author}{A.~Parrish}, \bibinfo{author}{A.~Mueller}, \bibinfo{author}{A.~Prabhu}, \bibinfo{author}{E.~McLean}, \bibinfo{author}{X.~Shen}, \bibinfo{author}{J.~Cavanagh}, \bibinfo{author}{A.~G. Gritsevskiy}, \bibinfo{author}{D.~Kauffman}, \bibinfo{author}{A.~T. Kirtland}, \bibinfo{author}{Z.~Zhou}, \bibinfo{author}{Y.~Zhang}, \bibinfo{author}{S.~Huang}, \bibinfo{author}{D.~Wurgaft}, \bibinfo{author}{M.~Weiss}, \bibinfo{author}{A.~Ross}, \bibinfo{author}{G.~Recchia}, \bibinfo{author}{A.~Liu}, \bibinfo{author}{J.~Liu}, \bibinfo{author}{T.~Tseng}, \bibinfo{author}{T.~Korbak}, \bibinfo{author}{N.~Kim}, \bibinfo{author}{S.~R. Bowman}, \bibinfo{author}{E.~Perez},
\newblock \bibinfo{title}{Inverse scaling: When bigger isn't better},
\newblock \bibinfo{journal}{Transactions on Machine Learning Research}  (\bibinfo{year}{2023}). \URLprefix \url{https://openreview.net/forum?id=DwgRm72GQF}, \bibinfo{note}{featured Certification}.
\bibitem[{Wei et~al.(2023)Wei, Kim, Tay, and Le}]{wei2023inverse}
\bibinfo{author}{J.~Wei}, \bibinfo{author}{N.~Kim}, \bibinfo{author}{Y.~Tay}, \bibinfo{author}{Q.~Le},
\newblock \bibinfo{title}{Inverse scaling can become u-shaped},
\newblock in: \bibinfo{booktitle}{Proceedings of the 2023 Conference on Empirical Methods in Natural Language Processing}, \bibinfo{year}{2023}, pp. \bibinfo{pages}{15580--15591}.
\bibitem[{Lourie et~al.(2025)Lourie, Hu, and Cho}]{lourie2025scaling}
\bibinfo{author}{N.~Lourie}, \bibinfo{author}{M.~Y. Hu}, \bibinfo{author}{K.~Cho},
\newblock \bibinfo{title}{Scaling laws are unreliable for downstream tasks: A reality check},
\newblock in: \bibinfo{editor}{C.~Christodoulopoulos}, \bibinfo{editor}{T.~Chakraborty}, \bibinfo{editor}{C.~Rose}, \bibinfo{editor}{V.~Peng} (Eds.), \bibinfo{booktitle}{Findings of the Association for Computational Linguistics: EMNLP 2025}, \bibinfo{publisher}{Association for Computational Linguistics}, \bibinfo{address}{Suzhou, China}, \bibinfo{year}{2025}, pp. \bibinfo{pages}{16167--16180}. \URLprefix \url{https://aclanthology.org/2025.findings-emnlp.877/}. \DOIprefix\doi{10.18653/v1/2025.findings-emnlp.877}.
\bibitem[{Fedus et~al.(2022)Fedus, Zoph, and Shazeer}]{fedus2022switch}
\bibinfo{author}{W.~Fedus}, \bibinfo{author}{B.~Zoph}, \bibinfo{author}{N.~Shazeer},
\newblock \bibinfo{title}{Switch transformers: Scaling to trillion parameter models with simple and efficient sparsity},
\newblock \bibinfo{journal}{Journal of Machine Learning Research} \bibinfo{volume}{23} (\bibinfo{year}{2022}) \bibinfo{pages}{1--39}.
\bibitem[{Giglou et~al.(2026)Giglou, D'Souza, Aioanei, Mihindukulasooriya, and Auer}]{giglou2026ontolearnermodularpythonlibrary}
\bibinfo{author}{H.~B. Giglou}, \bibinfo{author}{J.~D'Souza}, \bibinfo{author}{A.~Aioanei}, \bibinfo{author}{N.~Mihindukulasooriya}, \bibinfo{author}{S.~Auer}, \bibinfo{title}{Ontolearner: A modular python library for ontology learning with large language models}, \bibinfo{year}{2026}. \URLprefix \url{https://arxiv.org/abs/2607.01977}. \href{http://arxiv.org/abs/2607.01977}{{\tt arXiv:2607.01977}}.
\bibitem[{{Qwen Team}(2026{\natexlab{a}})}]{qwen3.5}
\bibinfo{author}{{Qwen Team}}, \bibinfo{title}{{Qwen3.5}: Towards native multimodal agents}, \bibinfo{year}{2026}{\natexlab{a}}. \URLprefix \url{https://qwen.ai/blog?id=qwen3.5}.
\bibitem[{{Qwen Team}(2026{\natexlab{b}})}]{qwen3.6-27b}
\bibinfo{author}{{Qwen Team}}, \bibinfo{title}{{Qwen3.6-27B}: Flagship-level coding in a {27B} dense model}, \bibinfo{year}{2026}{\natexlab{b}}. \URLprefix \url{https://qwen.ai/blog?id=qwen3.6-27b}.
\bibitem[{{Qwen Team}(2026{\natexlab{c}})}]{qwen36_35b_a3b}
\bibinfo{author}{{Qwen Team}}, \bibinfo{title}{{Qwen3.6-35B-A3B}: Agentic coding power, now open to all}, \bibinfo{year}{2026}{\natexlab{c}}. \URLprefix \url{https://qwen.ai/blog?id=qwen3.6-35b-a3b}.
\bibitem[{{OpenAI}(2026{\natexlab{a}})}]{openai_gpt55_api_2026}
\bibinfo{author}{{OpenAI}}, \bibinfo{title}{Gpt-5.5 model}, \bibinfo{year}{2026}{\natexlab{a}}. \URLprefix \url{https://developers.openai.com/api/docs/models/gpt-5.5}.
\bibitem[{{OpenAI}(2026{\natexlab{b}})}]{openai_gpt56_luna_2026}
\bibinfo{author}{{OpenAI}}, \bibinfo{title}{Gpt-5.6 luna model}, \bibinfo{year}{2026}{\natexlab{b}}. \URLprefix \url{https://developers.openai.com/api/docs/models/gpt-5.6-luna}.
\bibitem[{{OpenAI}(2026{\natexlab{c}})}]{openai_gpt56_terra_2026}
\bibinfo{author}{{OpenAI}}, \bibinfo{title}{Gpt-5.6 terra model}, \bibinfo{year}{2026}{\natexlab{c}}. \URLprefix \url{https://developers.openai.com/api/docs/models/gpt-5.6-terra}.
\bibitem[{{OpenAI}(2026{\natexlab{d}})}]{openai_gpt56_sol_2026}
\bibinfo{author}{{OpenAI}}, \bibinfo{title}{Gpt-5.6 sol model}, \bibinfo{year}{2026}{\natexlab{d}}. \URLprefix \url{https://developers.openai.com/api/docs/models/gpt-5.6-sol}.
\bibitem[{Ivanova and Terzieva(2026)}]{ivanova2026ontology}
\bibinfo{author}{T.~Ivanova}, \bibinfo{author}{V.~Terzieva},
\newblock \bibinfo{title}{Ontology learning in educational systems},
\newblock \bibinfo{journal}{Information} \bibinfo{volume}{17} (\bibinfo{year}{2026}). \URLprefix \url{https://www.mdpi.com/2078-2489/17/2/147}. \DOIprefix\doi{10.3390/info17020147}.
\bibitem[{Lippolis et~al.(2025{\natexlab{a}})Lippolis, Saeedizade, Keskis{\"a}rkk{\"a}, Zuppiroli, Ceriani, Gangemi, Blomqvist, and Nuzzolese}]{lippolis2025ontologygeneration}
\bibinfo{author}{A.~S. Lippolis}, \bibinfo{author}{M.~J. Saeedizade}, \bibinfo{author}{R.~Keskis{\"a}rkk{\"a}}, \bibinfo{author}{S.~Zuppiroli}, \bibinfo{author}{M.~Ceriani}, \bibinfo{author}{A.~Gangemi}, \bibinfo{author}{E.~Blomqvist}, \bibinfo{author}{A.~G. Nuzzolese},
\newblock \bibinfo{title}{Ontology generation using large language models},
\newblock in: \bibinfo{booktitle}{European Semantic Web Conference}, \bibinfo{organization}{Springer}, \bibinfo{year}{2025}{\natexlab{a}}, pp. \bibinfo{pages}{321--341}.
\bibitem[{Lippolis et~al.(2025{\natexlab{b}})Lippolis, Saeedizade, Keskisarkka, Gangemi, Blomqvist, and Nuzzolese}]{lippolis2025assessing}
\bibinfo{author}{A.~S. Lippolis}, \bibinfo{author}{M.~J. Saeedizade}, \bibinfo{author}{R.~Keskisarkka}, \bibinfo{author}{A.~Gangemi}, \bibinfo{author}{E.~Blomqvist}, \bibinfo{author}{A.~G. Nuzzolese},
\newblock \bibinfo{title}{Assessing the capability of large language models for domain-specific ontology generation},
\newblock \bibinfo{journal}{arXiv preprint arXiv:2504.17402}  (\bibinfo{year}{2025}{\natexlab{b}}).
\bibitem[{Teplyi and Dosyn(2025)}]{teplyi2025promptguided}
\bibinfo{author}{Y.~Teplyi}, \bibinfo{author}{D.~Dosyn},
\newblock \bibinfo{title}{Prompt-guided llm agent for end-to-end ontology learning},
\newblock \bibinfo{journal}{Computer Engineering} \bibinfo{volume}{22} (\bibinfo{year}{2025}) \bibinfo{pages}{35--47}.
\bibitem[{Fathallah et~al.(2026)Fathallah, Das, De~Giorgis, Poltronieri, Haase, Kovriguina, Mero{\~n}o-Pe{\~n}uela, Simperl, Staab, and Algergawy}]{fathallah2026extended}
\bibinfo{author}{N.~Fathallah}, \bibinfo{author}{A.~Das}, \bibinfo{author}{S.~De~Giorgis}, \bibinfo{author}{A.~Poltronieri}, \bibinfo{author}{P.~Haase}, \bibinfo{author}{L.~Kovriguina}, \bibinfo{author}{A.~Mero{\~n}o-Pe{\~n}uela}, \bibinfo{author}{E.~Simperl}, \bibinfo{author}{S.~Staab}, \bibinfo{author}{A.~Algergawy},
\newblock \bibinfo{title}{Extended neon-gpt: Advancing llm-powered ontology learning through ontology reuse and automated verification},
\newblock \bibinfo{journal}{Semantic Web} \bibinfo{volume}{17} (\bibinfo{year}{2026}) \bibinfo{pages}{22104968261453138}.
\bibitem[{Zhang et~al.(2025)Zhang, Dalal, Martin, Gadusu, and McGinty}]{zhang2025olive}
\bibinfo{author}{Y.~Zhang}, \bibinfo{author}{A.~S. Dalal}, \bibinfo{author}{C.~Martin}, \bibinfo{author}{S.~R. Gadusu}, \bibinfo{author}{H.~K. McGinty},
\newblock \bibinfo{title}{Olive: Ontology learning with integrated vector embeddings},
\newblock \bibinfo{journal}{Applied Ontology} \bibinfo{volume}{20} (\bibinfo{year}{2025}) \bibinfo{pages}{36--53}.
\bibitem[{Doumanas et~al.(2025)Doumanas, Soularidis, Spiliotopoulos, Vassilakis, and Kotis}]{doumanas2025finetuning}
\bibinfo{author}{D.~Doumanas}, \bibinfo{author}{A.~Soularidis}, \bibinfo{author}{D.~Spiliotopoulos}, \bibinfo{author}{C.~Vassilakis}, \bibinfo{author}{K.~Kotis},
\newblock \bibinfo{title}{Fine-tuning large language models for ontology engineering: A comparative analysis of gpt-4 and mistral},
\newblock \bibinfo{journal}{Applied Sciences} \bibinfo{volume}{15} (\bibinfo{year}{2025}). \URLprefix \url{https://www.mdpi.com/2076-3417/15/4/2146}. \DOIprefix\doi{10.3390/app15042146}.
\bibitem[{Liu et~al.(2026)Liu, Li, He, Ma, Wu, Yilmaz, Xia, Li, Tan, Fuh et~al.}]{liu2026prompt}
\bibinfo{author}{X.~Liu}, \bibinfo{author}{Z.~Li}, \bibinfo{author}{M.~He}, \bibinfo{author}{Z.~Ma}, \bibinfo{author}{X.~Wu}, \bibinfo{author}{G.~Yilmaz}, \bibinfo{author}{Y.~Xia}, \bibinfo{author}{B.~Li}, \bibinfo{author}{H.~Tan}, \bibinfo{author}{J.~Y.~H. Fuh}, et~al.,
\newblock \bibinfo{title}{From prompt to graph: Comparing llm-based information extraction strategies in domain-specific ontology development},
\newblock \bibinfo{journal}{arXiv preprint arXiv:2602.00699}  (\bibinfo{year}{2026}).
\bibitem[{Cappelli and Di~Marzo~Serugendo(2025)}]{cappelli2025methodological}
\bibinfo{author}{M.~A. Cappelli}, \bibinfo{author}{G.~Di~Marzo~Serugendo},
\newblock \bibinfo{title}{Methodological exploration of ontology generation with a dedicated large language model},
\newblock \bibinfo{journal}{Electronics} \bibinfo{volume}{14} (\bibinfo{year}{2025}) \bibinfo{pages}{2863}.
\bibitem[{da~Cruz et~al.(2025)da~Cruz, Tavares, and Belo}]{dacruz2025ontology}
\bibinfo{author}{T.~da~Cruz}, \bibinfo{author}{B.~Tavares}, \bibinfo{author}{F.~Belo},
\newblock \bibinfo{title}{Ontology learning and knowledge graph construction: A comparison of approaches and their impact on rag performance},
\newblock \bibinfo{journal}{arXiv preprint arXiv:2511.05991}  (\bibinfo{year}{2025}).
\bibitem[{Lippolis et~al.(2026)Lippolis, Saeedizade, Schmid, Blattner, Keskis{\"a}rkk{\"a}, Gangemi, Blomqvist, and Nuzzolese}]{lippolis2026ontoextend}
\bibinfo{author}{A.~S. Lippolis}, \bibinfo{author}{M.~J. Saeedizade}, \bibinfo{author}{S.~Schmid}, \bibinfo{author}{S.~Blattner}, \bibinfo{author}{R.~Keskis{\"a}rkk{\"a}}, \bibinfo{author}{A.~Gangemi}, \bibinfo{author}{E.~Blomqvist}, \bibinfo{author}{A.~G. Nuzzolese},
\newblock \bibinfo{title}{Ontoextend: A framework for requirement-driven and scalable ontology extension with llms},
\newblock \bibinfo{journal}{arXiv preprint arXiv:2607.17963}  (\bibinfo{year}{2026}).
\bibitem[{Cherian et~al.(2025)Cherian, Wang, and Zahid}]{cherian2025custom}
\bibinfo{author}{T.~Cherian}, \bibinfo{author}{W.~Y.~C. Wang}, \bibinfo{author}{A.~Zahid},
\newblock \bibinfo{title}{Custom fhir extensions with ontology learning models for improving data integration and interoperability in healthcare management systems},
\newblock in: \bibinfo{booktitle}{Proceedings of the 2025 9th International Conference on Medical and Health Informatics}, \bibinfo{year}{2025}, pp. \bibinfo{pages}{132--136}.
\bibitem[{Zhang et~al.(2025)Zhang, Li, Long, Zhang, Lin, Yang, Xie, Yang, Liu, Lin, Huang, and Zhou}]{qwen3embedding}
\bibinfo{author}{Y.~Zhang}, \bibinfo{author}{M.~Li}, \bibinfo{author}{D.~Long}, \bibinfo{author}{X.~Zhang}, \bibinfo{author}{H.~Lin}, \bibinfo{author}{B.~Yang}, \bibinfo{author}{P.~Xie}, \bibinfo{author}{A.~Yang}, \bibinfo{author}{D.~Liu}, \bibinfo{author}{J.~Lin}, \bibinfo{author}{F.~Huang}, \bibinfo{author}{J.~Zhou},
\newblock \bibinfo{title}{Qwen3 embedding: Advancing text embedding and reranking through foundation models},
\newblock \bibinfo{journal}{arXiv preprint arXiv:2506.05176}  (\bibinfo{year}{2025}).
\bibitem[{Beygi~Nasrabadi et~al.(2025)Beygi~Nasrabadi, Norouzi, Hubaiev, Waitelonis, and Sack}]{beygi2025nfdi}
\bibinfo{author}{H.~Beygi~Nasrabadi}, \bibinfo{author}{E.~Norouzi}, \bibinfo{author}{K.~Hubaiev}, \bibinfo{author}{J.~Waitelonis}, \bibinfo{author}{H.~Sack},
\newblock \bibinfo{title}{Nfdi matwerk ontology (mwo): A bfo-compliant ontology for research data management in materials science and engineering},
\newblock \bibinfo{journal}{Advanced Engineering Materials}  (\bibinfo{year}{2025}) \bibinfo{pages}{e202502331}.
\bibitem[{Rajamohan et~al.(2025)Rajamohan, Bradley, Tran, Gordon, Caldwell, Mehdi, Ponon, Tran, Dernek, Kaltenbaugh et~al.}]{rajamohan2025materials}
\bibinfo{author}{B.~P. Rajamohan}, \bibinfo{author}{A.~C.~H. Bradley}, \bibinfo{author}{V.~D. Tran}, \bibinfo{author}{J.~E. Gordon}, \bibinfo{author}{H.~W. Caldwell}, \bibinfo{author}{R.~Mehdi}, \bibinfo{author}{G.~Ponon}, \bibinfo{author}{Q.~D. Tran}, \bibinfo{author}{O.~Dernek}, \bibinfo{author}{J.~Kaltenbaugh}, et~al.,
\newblock \bibinfo{title}{Materials data science ontology (mds-onto): Unifying domain knowledge in materials and applied data science},
\newblock \bibinfo{journal}{Scientific Data} \bibinfo{volume}{12} (\bibinfo{year}{2025}) \bibinfo{pages}{628}.
\bibitem[{Bandrowski et~al.(2016)Bandrowski, Brinkman, Brochhausen, Brush, Bug, Chibucos, Clancy, Courtot, Derom, Dumontier et~al.}]{OBI}
\bibinfo{author}{A.~Bandrowski}, \bibinfo{author}{R.~Brinkman}, \bibinfo{author}{M.~Brochhausen}, \bibinfo{author}{M.~H. Brush}, \bibinfo{author}{B.~Bug}, \bibinfo{author}{M.~C. Chibucos}, \bibinfo{author}{K.~Clancy}, \bibinfo{author}{M.~Courtot}, \bibinfo{author}{D.~Derom}, \bibinfo{author}{M.~Dumontier}, et~al.,
\newblock \bibinfo{title}{The ontology for biomedical investigations},
\newblock \bibinfo{journal}{PloS one} \bibinfo{volume}{11} (\bibinfo{year}{2016}) \bibinfo{pages}{e0154556}.
\bibitem[{Hastings et~al.(2011)Hastings, Ceusters, Smith, and Mulligan}]{hastings2011dispositions}
\bibinfo{author}{J.~Hastings}, \bibinfo{author}{W.~Ceusters}, \bibinfo{author}{B.~Smith}, \bibinfo{author}{K.~Mulligan},
\newblock \bibinfo{title}{Dispositions and processes in the emotion ontology}  (\bibinfo{year}{2011}).
\bibitem[{Hastings et~al.(2012)Hastings, Ceusters, Mulligan, and Smith}]{hastings2012annotating}
\bibinfo{author}{J.~Hastings}, \bibinfo{author}{W.~Ceusters}, \bibinfo{author}{K.~Mulligan}, \bibinfo{author}{B.~Smith},
\newblock \bibinfo{title}{Annotating affective neuroscience data with the emotion ontology}  (\bibinfo{year}{2012}).

\end{thebibliography}

\appendix

\end{document}